\documentclass{article}

\usepackage{PRIMEarxiv}
\usepackage{amsmath}        
\usepackage[T1]{fontenc}    
\usepackage{hyperref}       
\usepackage{url}            
\usepackage{booktabs}       
\usepackage{amsfonts}       
\usepackage{nicefrac}       
\usepackage{microtype}      
\usepackage{lipsum}
\usepackage{fancyhdr}       
\usepackage{graphicx}       
\graphicspath{{media/}}     

\usepackage{hyperref}
\DeclareMathOperator{\sinc}{sinc}
\title{A survey on recently proposed activation functions for Deep Learning
}

\author{
  Murilo Gustineli \\
  \texttt{murilogustineli@gmail.com}}

\begin{document}
\maketitle

\begin{abstract}

Artificial neural networks (ANN), typically referred to as neural networks, are a class of Machine Learning algorithms and have achieved widespread success, having been inspired by the biological structure of the human brain. Neural networks are inherently powerful due to their ability to learn complex function approximations from data. This generalization ability has been able to impact multidisciplinary areas involving image recognition, speech recognition, natural language processing, and others. Activation functions are a crucial sub-component of neural networks. They define the output of a node in the network given a set of inputs. This survey discusses the main concepts of activation functions in neural networks, including; a brief introduction to deep neural networks, a summary of what are activation functions and how they are used in neural networks, their most common properties, the different types of activation functions, some of the challenges, limitations, and alternative solutions faced by activation functions,  concluding with the final remarks.
 
\end{abstract}

\section{Introduction}
Modern artificial neural network architecture draws inspiration from the neurological structure of the brain, which is a complex mesh of input-output wirings between neurons that cascade decision signals across a network. A deep neural network (DNN) is an artificial neural network with multiple layers between the input and output layers \cite{schmidhuber2015deep}. The layers between the input and output layers are called hidden layers. A network with only one hidden layer is called a “shallow” neural network. Deep learning is a subset of machine learning where neural networks learn from large datasets. The adjective "deep" in deep learning refers to the use of multiple layers in the network. Neurons, inputs, weights, biases, and functions are essential components of neural networks \cite{lecun2015deep}. These components work conjointly similarly to the human brain and can be trained just like any other machine learning algorithm. The figure below represents a multilayer neural network, consisting of three neurons in the input layer, two hidden layers with five neurons each, and two neurons in the output layer.

\begin{figure}[ht]
\centering
\includegraphics[width=0.45\textwidth]{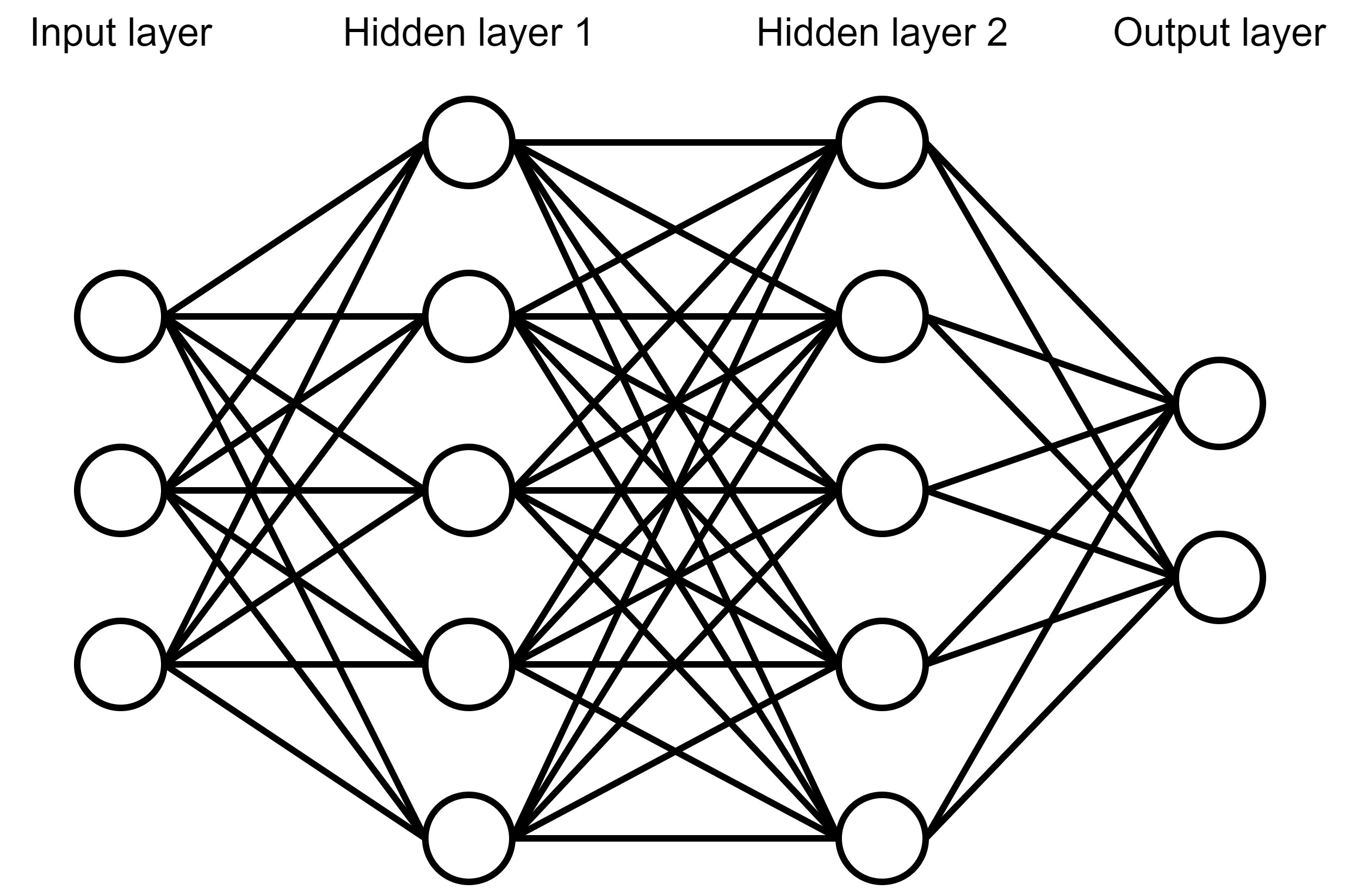}
\caption{\label{fig:neural_nets}Structure of a Multilayer Neural Network}
\end{figure}

\section{Background}

An activation function is a nonlinear function used to train a neural network by defining the output of a neuron given a set of inputs \cite{lecun1989backpropagation}. This function is used in every neuron of the network to help the network learn the complex patterns in the data, allowing it to make predictions. Moreover, the purpose of activation functions is to introduce nonlinearities into the neural network.

An artificial neural network is composed of a large number of connected nodes known as perceptrons or neurons. A perceptron (figure below) receives real-valued numerical inputs of a feature (x) and multiplies each input with its associated weights (w) and bias (b) to compute the net input, which is the sum of all connected neurons. The result of net input is passed on to the activation function that mathematically transforms the output into the range of {-1, 1} classified by the hyperplane at the origin. These distinct planes translate to the predicted class label of the input data points. This output is used during the learning phase to calculate the prediction error and to update the weights and bias unit.

\begin{figure}[ht]
\centering
\includegraphics[width=0.7\textwidth]{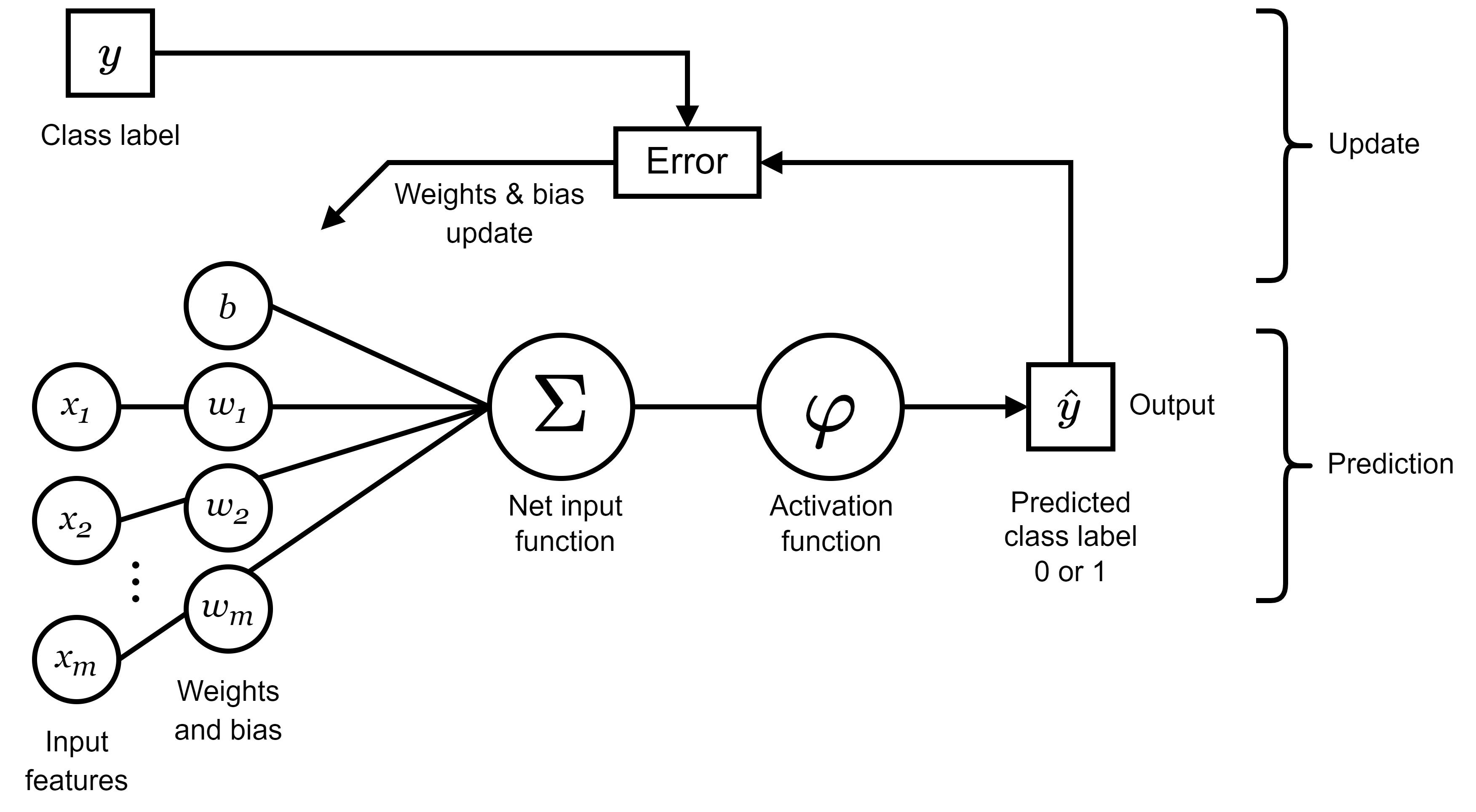}
\caption{\label{fig:activation_function}The Perceptron: Example}
\end{figure}

The first component of the perceptron is the input matrix. The input is a $m$ dimensional matrix $x\ =\ [ x_{1} \ x_{2} \ ...\ x_{m}]$ where each $x_{i}$ represents a single data point. For example, if the input matrix of the neural network is an image, a single data point  would be a pixel of that image. The dimension of the input matrix will be dependent on the size of the image. Hence, a black and white image of 28x28 pixels will have an input matrix of dimension $28\times 28$ where $x_{m} = 784$.
The second component of the perceptron is the weights matrix. The weights matrix has the same dimension as the input matrix, thus, $w\ =\ [ b\ w_{1} \ w_{2} \ ...\ w_{m}]$, where each $w_{i}$ represents the weights associated with its corresponding inputs, and the $b$ represents the bias unit. The summation of the inner product of the input matrix and weight matrix is passed on to the activation function that produces an output between 0 and 1. Note, the representation of a perceptron may vary depending on the structure of the neural network.

\section{Activation function properties}

Activation functions share common properties. Some of the most important properties are nonlinearity, differentiability, continuous, bounded, and zero-centering \cite{datta2020survey}.
All activation functions are nonlinear functions, meaning, the graph of the function is different from a line where a change of the input is not proportional to the change of the output.
Differentiability is related to the derivative of the function. A differentiable function of one real variable is one whose derivative occurs at each point in its domain. The gradient of the loss function is calculated during backpropagation using the gradient descent method \cite{lecun1989backpropagation}. As a result, the activation function must be differentiable with respect to its input.
A continuous function is one in which a continuous variation of the argument causes a continuous variation of the function's value. This means that there are no abrupt changes in value, which are referred to as discontinuities.
Bounded functions are limited by some form of boundary or restriction. A bounded function's range has both a lower and upper bound. This is relevant for neural networks because the activation function is responsible for keeping the output values within a certain range, otherwise, the values may exceed reasonable values.
When a function's range contains both positive and negative values, it is said to be zero-centered. If a function is not zero-centered, as the sigmoid function, the output of a layer is always shifted to either positive or negative values. As a result, the weight matrix requires more updates to be adequately trained, increasing the number of epochs needed to train the network. This is why the zero-centered property is useful, even if it isn't required.
The computational cost of an activation function is defined as the time required to generate the activation function's output when given input. The computational cost of the gradient is equally important as it is calculated during backpropagation when the weights are updated. Low computational cost requires less time for a neural network to be trained, while the opposite is also true. Hence, the search to find a function with lower computational cost has been highly  desired in the machine learning research field.

\section{Challenges faced by activation functions}

The two major challenges faced by widely used activation functions are the vanishing gradient problem and the dead neuron problem. This section will cover both issues.

\subsection{Vanishing gradient problem}

This problem happens when the values of the gradient get closer to zero as the backpropagation goes deeper into the network, making the weights saturated and not updated properly. As a result, the loss stops decreasing and the network does not get trained properly. This problem is termed the vanishing gradient problem. The neurons whose weights are not properly updated are referred to as saturated neurons.

\subsection{Dead neuron problem}

As previously discussed, the output value of activation functions ranges between 0 and 1. When the value is close to zero, it forces the corresponding neurons to be inactive, thus, not contributing to the final output. Moreover, the weights may be updated in such a way that the weighted sum of a large portion of the network is forced to zero. This scenario may force a large portion of the input to be deactivated, resulting in an unrecoverable problem during the network performance. Hence, these neurons that have been forcefully deactivated are known as "dead neurons," and the problem is known as the "dead neuron problem".

\section{Different types of activation functions}

The most commonly used activation functions in recent years are the Sigmoid, Tanh, ReLU, LReLU, and PReLU.

\subsection{Sigmoid}

The sigmoid function, often called the logistic sigmoid function, is one of  the most commonly known functions used in feedforward neural networks today. This is primarily due to its nonlinearity and the simplicity of the derivative, which is relatively computationally inexpensive. AF sigmoid function is a bounded differentiable real function that is defined for all real input values and that has positive derivatives. The sigmoid function is defined as

$$
f( x) \ =\ \frac{1}{1+e^{-x}}
$$

Further, a sigmoidal function is assumed to rapidly approach a fixed finite upper limit asymptotically as its argument gets large, and to rapidly approach a fixed finite lower limit asymptotically as its argument gets small. The central portion of the sigmoid (whether it is near 0 or displaced) is assumed to be roughly linear \cite{han1995influence}.
The sigmoid function contains an exponential term as it can be seen from the function definition. Exponential functions, such as the sigmoid function, have high computation cost.Although the function is computationally expensive, its gradient is not. The gradient can be computed by using the formula

$$
f\prime ( x) \ =\ f( x)( 1-f( x))
$$

A major drawback is that the sigmoid function is bound in a range between 0 and 1. Thus, it always produces a non-negative value as output. The sigmoid function binds a large range of inputs to a small range between 0 and 1. Therefore, a large change to the input value leads to a small change to the output value, resulting into small gradient values as well. Because of the small gradient values, it may be prone to suffering from the vanishing gradient problem. The hyperbolic tangent, or tanh function, was created to combine the advantages of the sigmoid function with its zero-centered nature.

\subsection{Tanh}

The hyperbolic tangent, or tanh function became more popular than the sigmoid function because in most cases, it gives better training performance for multi-layer neural networks. The tanh function inherits all the valuable properties of the sigmoid function. The tanh function is defined as

$$
f( x) \ =\ \frac{1-e^{-x}}{1+e^{-x}}
$$

The tanh function is continuous, differentiable and bounded, and ranges between -1 and 1. Therefore, the range of possible outputs expanded, including negative, positive, and zero outputs. Moreover, the tanh function is zero-centered, hence, reducing the number of epochs needed to train the network as compared to the sigmoid function. The zero-centered property is one of the main advantages provided by the tanh function, thereby helping the backpropagation process. Both the sigmoid and tanh functions are computationally expensive because they are exponential functions. 
The tanh function, in a similar way to sigmoid, binds a large range of input to a small range between -1 and 1. Thus,  a large change to the input value leads to a small change to the output value. This results in close to zero gradient values. Because the gradient values may get close to zero, tanh suffers vanishing gradient problems. The vanishing gradient problem prompted more research into activation functions, which led to the development of ReLU.

\subsection{ReLU}

Since it’s proposal \cite{hahnloser2000digital}, the rectified linear unit function (ReLU) has been widely used in neural networks because of its efficient properties. The ReLU function is defined as

$$
f( x) \ =\ max( 0,\ x)
$$

where x is the input to the activation function. The ReLU function is continuous, not-bounded and not zero-centered. Different than the sigmoid and tanh function, ReLU is not exponential, thus, it has low computational cost as it forces negative values to zero. This feature makes the ReLU function a better candidate to be used in neural networks as it provides better performance and generalization when compared to the sigmoid and tanh functions.
Negative inputs passed on to the ReLU function are evaluated to a zero output. As a consequence, negatively weighted neurons do not contribute to the overall performance of the network, suffering the previously seen dead neuron problem. A new variant of the ReLU, called LReLU, was introduced in an attempt to solve the dead neuron problem.

\subsection{LReLU}

The leaky ReLU (LReLU) function is continuous, not-bounded, zero-centered, and it has low computational cost. The LReLU is defined as

$$
f( x) \ =\ \begin{cases}
0.01x & \text{for} \ x\leq 0\\
x & \text{otherwise}
\end{cases}
$$

Unlike the ReLU, the LReLU function allows negative inputs to be passed on as outputs. For x input values smaller than 0, the left-hand derivative is 0.01, while the right-hand derivative is 1. For x input values greater than 0, the gradient is always 1, hence, the function does not suffer from the vanishing gradient problem. On the other hand, the gradient of negative outputs will always be 0.01, leading to a risk of potentially suffering from the vanishing gradient problem.

\subsection{PReLU}

To solve the vanishing gradient problem faced by the leaky ReLU function, the parametric ReLU (PReLU) function was introduced by He et al. \cite{he2015delving}. The PReLU function is defined as

$$
f( x) \ =\ \begin{cases}
ax & \text{for} \ x\leq 0\\
x & \text{otherwise}
\end{cases}
$$

where a is a learnable parameter and x is the input to the activation function. When a is 0.01, PReLU function is equal to LReLU, and when a = 0, PReLU function is equal to ReLU. As a result, PReLU can be generally used to express rectifier nonlinearities. The PReLU is non-bounded, continuous, and zero-centered. When x is less than zero, the function's gradient is a, and when x is greater than zero, the function's gradient is 1. There is no vanishing gradient problem in the positive part of the PReLU function, where the gradient is always 1. However, the gradient on the negative side is always a, which is often close to zero. It raises the possibility of a vanishing gradient problem.

\section{Alternative solutions for challenges faced by activation functions}

The adoption of the Rectified Linear Unit (ReLU) activation function to solve the vanishing gradient problem created by utilizing saturating activation functions has been a major discovery that made training deep networks possible. Many enhanced ReLU variations have been proposed since then. This section will cover the most recent breakthroughs of activation functions that attempt to solve some of the limitations faced by the novel ReLU function used in neural networks.

\subsection{Swish}

Reinforcement learning based search techniques have been used to discover novel activation functions that could be considered potential replacements for the highly used ReLU function, leading to the discovery of the Swish function \cite{ramachandran2017searching}. Swish has been found to outperform ReLU on deeper models across a variety of large datasets. On nearly all tasks, Swish matches or exceeds the baselines when compared to the ReLU and other novel activation functions. The Swish formula is defined as

$$
f( x) \ =\ x\times \text{sigmoid}( \beta x)
$$

Swish's simplicity and resemblance to ReLU allows practitioners to easily replace ReLUs with Swish units in any neural network by changing a single line of code. Swish is unbounded above and bounded below. Unlike ReLU, Swish is smooth and nonmonotonic. Thus, the non-monotonicity property of Swish is a key differentiator from the most common activation functions.Swish's smooth, continuous profile was critical in deep neural network architectures for better information propagation when compared to ReLU. The key differentiator between Swish and ReLU is the non-monotonic “bump” of Swish when x < 0.  The conducted study makes an empirical case for how much of the pre-activation signal falling in the -limit<x<0 (-limit is predefined) range is captured by this bump. Further, the bump can be tuned based on the chosen architecture according to its  use case. Because of this intuition, Swish is able to outperform other monotonic activation functions, making Swish a good candidate as a replacement for ReLU.

\subsection{Mish}

Mish is a self-regularized, smooth, continuous, non-monotonic activation function inspired by Swish's self-gating property. Mish has been proposed \cite{misra2019mish} as a replacement for novel activation functions, including Swish, which is a more robust activation function with significantly better results than ReLU. The Mish function is defined as

$$
f( x) \ =\ x\ \text{tanh}\left(\text{softplus}( x)\right)
$$

Mish tends to match or improve the performance of neural network architectures when compared to Swish, ReLU, and Leaky ReLU across various Computer Vision tasks. Even though Mishs’s design has been influenced by the work performed by Swish, it was discovered by systematic analysis and experimentation of the characteristics that made Swish so successful. Mish is bounded below and unbounded above, similar to Swish. According to the research, Mish consistently outperformed the aforementioned functions, as well as Swish and ReLU.

Mish uses the Self-Gating property, which involves multiplying the non-modulated input with the output of a nonlinear function of the input. Mish eliminates the necessary preconditions for the Dying ReLU problem by retaining a small quantity of negative information. This property aids in the betterment of network information flow. Mish is unbounded above, avoiding saturation and the vanishing gradient problem, which causes training to slow down. It's also advantageous to be bounded below because it results in stronger regularization effects. Unlike ReLU, Mish is continuously differentiable, which is preferable since it avoids singularities, thus, unwanted side effects when doing gradient-based optimization.

Mish’s smooth profile also plays a role in better gradient flow. Smoother output landscapes imply smooth loss landscapes, making optimization and generalization easier. Mish has a larger minima than ReLU and Swish, which increases generalization because the former contains several local minima. Furthermore, as compared to the networks equipped with ReLU and Swish, Mish had the lowest loss, proving the preconditioning effect of Mish on the loss surface.
Under most experimental conditions, Mish outperforms Swish, ReLU, and Leaky ReLU in terms of empirical data.

\subsection{GCU}

Papert and Minsky \cite{mycielski1972marvin} were the first to point out that a single neuron can't learn the XOR function since a single hyperplane (in this example, a line) cannot separate the output classes for this function formulation. Moreover, the XOR problem needs at least 3 neurons to be solved. As an attempt to solve this problem a new activation function, Growing Cosine Unit (GCU) has been proposed \cite{noel2021growing} with the advantages of using oscillatory activation functions to improve gradient flow and alleviate the vanishing gradient problem. The GCU function is defined as

$$
C( z) \ =\ z\times \text{cos} \ z
$$

Because of their biological plausibility, most activation functions utilized today are non-oscillatory and monotonically growing. Oscillatory activation functions can improve gradient flow and reduce network size. Oscillatory activation functions are shown to have the following advantages; alleviate the vanishing gradient problem as they have non-zero derivatives throughout their domain except at isolated points, improve performance for compact network architectures, and are computationally cheaper than the state-of-the-art Swish and Mish activation functions.

As a result, oscillatory activation functions allow neurons to alter categorization (output sign) inside the interior of neuronal hyperplane positive and negative half-spaces, allowing complex decisions to be made with fewer neurons. Furthermore, the GCU oscillatory function presents a single neuron solution to the XOR problem. It has been found that a composite network trained with GCU outperforms standalone Sigmoids, Swish, Mish, and ReLU networks on a variety of architectures and benchmarks. The GCU function is computationally cheaper than Swish and Mish. The GCU activation also reduces training time and allows classification problems to be solved with smaller networks.

\section{Biologically inspired oscillating activation functions}

Recent research has discovered that biological neurons in the second and third layer of the human cortex have oscillating activation properties and are capable of individually learning the XOR function. The presence of oscillating activation functions in biological neural neurons may explain the difference in performance between biological and artificial neural networks. It has been demonstrated that oscillatory activation functions outperform popular activation functions on many tasks \cite{noel2021growing}. This paper proposes 4 new biologically inspired oscillating activation functions (Shifted Quadratic Unit (SQU), Non-Monotonic Cubic (NCU), Shifted Sinc Unit (SSU), and Decaying Sine Unit (DSU)) that enable single neurons to learn the XOR problem without manual feature engineering. The oscillatory activation functions also outperform popular activation functions on many benchmark tasks, such as solving classification problems with fewer neurons and reducing training time.

The vanishing gradient problem can be greatly reduced by using activations that have larger derivatives for a wider range of inputs, such as the ReLU, LReLU, PReLU, Swish, and Mish activation functions \cite{apicella2021survey}. These novel activation functions are unbounded, perform better in deep networks, and have derivative values near to one or greater for all positive values \cite{he2015delving}. However, the vast majority of activation functions in neural networks are monotonic or nearly monotonic functions with a single zero at the origin.
Albert Gidon et al., \cite{gidon2020dendritic} discovered a new type of neuron in the human cortex capable of learning the XOR function on its own (a task which is impossible with single neurons using sigmoidal, ReLU, LReLU, PReLU, Swish, and Mish activations).
To differentiate the classes in the XOR dataset, two hyperplanes are required, which necessitates the use of activation functions with multiple zeros, as described in \cite{lotfi2014novel}. Given that the output of biological neurons increases for small input values, it is evident that the output must ultimately decrease to zero if a biological neuron is capable of learning the XOR function. Consequently, although conventional activation functions need a 3 neuron network with 2 hidden and 1 output layer to learn the XOR function, the XOR function may be learned with a single neuron using oscillatory activation, such as Growing Cosine Unit (GCU).

It has been demonstrated that oscillatory activation functions outperform popular activation functions on many tasks \cite{noel2021growing} \cite{noel2021biologically}. In this paper 4 new oscillatory activation functions that enable individual artificial neurons to learn the XOR function like biological neurons are proposed. The oscillatory activation functions also outperform popular activation functions on many benchmark tasks, such as solving classification problems with fewer neurons and reduced training time.

The four oscillatory functions presented in the paper are defined as the following:
\begin{itemize}
\item Non-Monotonic Cubic Unit (NCU): $f( z) \ =\ z\ -\ z^{3}$

\item Shifted Quadratic Unit (SQU): $f( z) \ =\ z^{2} + z$

\item Decaying Sine Unit (DSU): $f(z) = \frac{\pi}{2}(\sinc{(z-\pi)}-\sinc{(z+\pi)})$

\item Shifted Sinc Unit (SSU): $f( z) \ =\ \pi \ \text{sinc}( z\ -\ \pi )$
\end{itemize}

The results from the study indicate the oscillatory activation functions like NCU, SQU, DSU, and SSU are converging at 20 epochs while maintaining the best possible accuracy at 4 convolution layers. They maintain accuracy over 68 percent with 3 convolution layers. Furthermore, Quadratic (SQU) retains the accuracy mentioned above even with a reduced number of epochs. The results show that networks with oscillating activations can use fewer neurons, perform better, and train quicker than networks with more common activation functions. Finally, the findings in this research show that deep networks with oscillating activation functions may be able to overcome the performance gap between biological and artificial neural networks. Future study is needed to assess the performance of the novel oscillatory activation functions described in this research on additional benchmark issues and model designs.

\section{Conclusion}
Neural networks are intrinsically powerful due to their ability to learn complex function approximations from data. This generalization ability has been able to impact multidisciplinary areas involving image recognition, speech recognition, natural language processing, and others. Activation functions are a crucial sub-component of neural networks. They define the output of a node in the network given a set of inputs. This paper discusses popular  activation functions and provides a comparison on their properties. Recent work has proposed oscillating activation functions outperform popular activation functions on many benchmark tasks, such as solving classification problems with fewer neurons and reduced training time.

\bibliographystyle{unsrt}  
\bibliography{references}

\begin{thebibliography}{10}

\bibitem{schmidhuber2015deep}
J{\"u}rgen Schmidhuber.
\newblock Deep learning in neural networks: An overview.
\newblock {\em Neural networks}, 61:85--117, 2015.

\bibitem{lecun2015deep}
Yann LeCun, Yoshua Bengio, and Geoffrey Hinton.
\newblock Deep learning.
\newblock {\em nature}, 521(7553):436--444, 2015.

\bibitem{lecun1989backpropagation}
Yann LeCun, Bernhard Boser, John~S Denker, Donnie Henderson, Richard~E Howard,
  Wayne Hubbard, and Lawrence~D Jackel.
\newblock Backpropagation applied to handwritten zip code recognition.
\newblock {\em Neural computation}, 1(4):541--551, 1989.

\bibitem{datta2020survey}
Leonid Datta.
\newblock A survey on activation functions and their relation with xavier and
  he normal initialization.
\newblock {\em arXiv preprint arXiv:2004.06632}, 2020.

\bibitem{han1995influence}
Jun Han and Claudio Moraga.
\newblock The influence of the sigmoid function parameters on the speed of
  backpropagation learning.
\newblock In {\em International workshop on artificial neural networks}, pages
  195--201. Springer, 1995.

\bibitem{hahnloser2000digital}
Richard~HR Hahnloser, Rahul Sarpeshkar, Misha~A Mahowald, Rodney~J Douglas, and
  H~Sebastian Seung.
\newblock Digital selection and analogue amplification coexist in a
  cortex-inspired silicon circuit.
\newblock {\em nature}, 405(6789):947--951, 2000.

\bibitem{he2015delving}
Kaiming He, Xiangyu Zhang, Shaoqing Ren, and Jian Sun.
\newblock Delving deep into rectifiers: Surpassing human-level performance on
  imagenet classification.
\newblock In {\em Proceedings of the IEEE international conference on computer
  vision}, pages 1026--1034, 2015.

\bibitem{ramachandran2017searching}
Prajit Ramachandran, Barret Zoph, and Quoc~V Le.
\newblock Searching for activation functions.
\newblock {\em arXiv preprint arXiv:1710.05941}, 2017.

\bibitem{misra2019mish}
Diganta Misra.
\newblock Mish: A self regularized non-monotonic activation function.
\newblock {\em arXiv preprint arXiv:1908.08681}, 2019.

\bibitem{mycielski1972marvin}
Jan Mycielski.
\newblock Marvin minsky and seymour papert, perceptrons, an introduction to
  computational geometry.
\newblock {\em Bulletin of the American Mathematical Society}, 78(1):12--15,
  1972.

\bibitem{noel2021growing}
Mathew~Mithra Noel, Advait Trivedi, Praneet Dutta, et~al.
\newblock Growing cosine unit: A novel oscillatory activation function that can
  speedup training and reduce parameters in convolutional neural networks.
\newblock {\em arXiv preprint arXiv:2108.12943}, 2021.

\bibitem{apicella2021survey}
Andrea Apicella, Francesco Donnarumma, Francesco Isgr{\`o}, and Roberto
  Prevete.
\newblock A survey on modern trainable activation functions.
\newblock {\em Neural Networks}, 138:14--32, 2021.

\bibitem{gidon2020dendritic}
Albert Gidon, Timothy~Adam Zolnik, Pawel Fidzinski, Felix Bolduan, Athanasia
  Papoutsi, Panayiota Poirazi, Martin Holtkamp, Imre Vida, and Matthew~Evan
  Larkum.
\newblock Dendritic action potentials and computation in human layer 2/3
  cortical neurons.
\newblock {\em Science}, 367(6473):83--87, 2020.

\bibitem{lotfi2014novel}
Ehsan Lotfi and M-R Akbarzadeh-T.
\newblock A novel single neuron perceptron with universal approximation and xor
  computation properties.
\newblock {\em Computational intelligence and neuroscience}, 2014, 2014.

\bibitem{noel2021biologically}
Matthew~Mithra Noel, Shubham Bharadwaj, Venkataraman Muthiah-Nakarajan, Praneet
  Dutta, and Geraldine~Bessie Amali.
\newblock Biologically inspired oscillating activation functions can bridge the
  performance gap between biological and artificial neurons.
\newblock {\em arXiv preprint arXiv:2111.04020}, 2021.

\end{thebibliography}

\end{document}